\documentclass[acmlarge]{acmart}
\pdfoutput=1
\usepackage{booktabs} 
\usepackage{tabu}

\usepackage{subfig}
\usepackage{listings}

\usepackage[ruled]{algorithm2e} 

\SetAlFnt{\small}
\SetAlCapFnt{\small}
\SetAlCapNameFnt{\small}
\SetAlCapHSkip{0pt}
\IncMargin{-\parindent}

\acmYear{2017}
\acmMonth{10}

\setcopyright{rightsretained}

\begin{document}
\title{Semantic Web Today: From Oil Rigs to Panama Papers} 
\author{Rivindu Perera$^{*}$}
\author{Parma Nand}
\author{Boris Ba\v{c}i\'{c}}
\affiliation{%
  \institution{Auckland University of Technology}
  \country{New Zealand}}
\author{Wen-Hsin Yang}
\affiliation{%
  \institution{Industrial Technology Research Institute}
  \country{Taiwan}
}
\author{Kazuhiro Seki}
\affiliation{%
  \institution{Kobe University}
  \country{Japan}}
\author{Radek Burget}
\affiliation{%
  \institution{Brno university of technology}
  \country{Czech Republic}}

\begin{abstract}
The next leap on the internet has already started as Semantic Web. At its core, Semantic Web transforms the document oriented web to a data oriented web enriched with semantics embedded as metadata. This change in perspective towards the web offers numerous benefits for vast amount of data intensive industries that are bound to the web and its related applications. The industries are diverse as they range from Oil \& Gas exploration to the investigative journalism, and everything in between. This paper discusses eight different industries which currently reap the benefits of Semantic Web. The paper also offers a future outlook into Semantic Web applications and discusses the areas in which Semantic Web would play a key role in the future.
\end{abstract}

%
%


%
%


\thanks{
$^{*}$ Corresponding author: rivindu.perera@aut.ac.nz

The work reported in this paper is part of the RealText project funded by Auckland
University of Technology.}

\maketitle

\renewcommand{\shortauthors}{R. Perera et al.}

\section{Introduction}

The concept of Semantic Web is fast becoming a key instrument in diverse fields since its inception in 2001 through the seminal paper by \citet{Berners-lee2001}. Semantic Web extends the current web infrastructure by transforming the document web to a data web (Web of Data) which is machine readable. It also enables us to create links within and across the data as it operates at the data level and thus is more granular than the document level linking in the traditional web. Since the Semantic Web works at the data level with a well specified meaning, this approach makes it widely applicable throughout diverse domains with seamless integration with a variety of technologies. The two paragraphs below discuss case studies to illustrate the potential of Semantic Web in two different areas. 

A single offshore oil rig produces terabytes of data in a single day, making Oil and Gas industry one of the most data intensive industries on the earth today. According to Chevron, a leading oil and gas company, the current data load is over 6000 TB and growing at a rate of 80\% annually. \citep{Crompton2008}. Instead of the traditional document collections, Chevron started employing the Semantic Web to manage the excessive data load. The company has came up with number of tools and techniques, and among them the ontology-driven information integration case study \citep{Chum2009} clearly shows the flexibility and the scalability of the data model. It also reports a number of other benefits which support easy handling of the massive data load. 

On 16 April 2016, \textquotedbl{}John Doe\textquotedbl{}, a whistle-blower released 2.6 Tb of data (including 320,166 text documents, 2.15 million PDF files, and 4.8 million emails) which is commonly known as Panama Papers and describes how politicians, business entities, and other individuals evade taxes and other international sanctions. After exactly one month, on 17 May 2016, a leading Semantic Web company, Ontotext, transformed the Panama Papers into Linked Data form and introduced this new data collection as Linked Leaks dataset \citep{Kiryakov2016} which is entirely based on the Semantic Web technologies. One of the main advantage of Linked Leaks dataset is that it is easy to query and identify how different entities mentioned in the dataset interrelate with each other, which was very difficult to carry out with the original unstructured text documents.

The aforementioned two case studies illustrate the integration of Semantic Web into two diverse fields in order to represent and process information originally encoded as natural text. The objective of this paper is to present an overview of applications and other areas where Semantic Web has contributed to improve the workflow with its wide range of technology stack. 

The rest of the paper is structured as follows. The next section is devoted for an introduction to the Semantic Web. We describe the essence and the foundations of the Semantic Web concepts and interested readers can follow the references to refer to extended discussions. Section~\ref{sec:oil-and-gas-industry} to Section~\ref{sec:investigative-journalism} focus on eight different domains which have benefited widely from the Semantic Web technologies. We present an overview of cutting-edge research being carried out in these areas as well as some future research opportunities and remaining gaps. Section~\ref{sec:semantic-web-tomorrow} provides a brief overview of how the semantic web can contribute to other areas which are not investigated so far. We conclude the paper in Section~\ref{sec:conclusio} with an analysis on the presented information.

\section{Semantic Web: A Gentle Introduction}

This section provides a brief introduction into the Semantic Web technology stack which provides the foundation for a number of applications in a wide variety of industries.

The origin of the Semantic Web addresses a number of drawbacks of the traditional document oriented web. Firstly, the documents which contain unstructured text are indexed and retrieved using text matching algorithms which do not take semantics of the document into account. This results in irrelevant documents in the search result. Secondly, the websites that generate the content based on structured backend data thus the underlying data remains hidden, hence is not available for consumption directly by machines. The third and the most important factor is that current web does not incorporate semantics as metadata. For example, headings in a HTML page are merely special appearances of text and has no metadata that denotes how that information is related to the core of the document. 

Semantic Web addresses all of the above issues by providing the access to the data together with the metadata that enables a semantic view of the information contained in the web page. This is enabled by a range of technologies which specifies guidelines on how this data should be presented. 

\begin{figure}
	\includegraphics[scale=0.7]{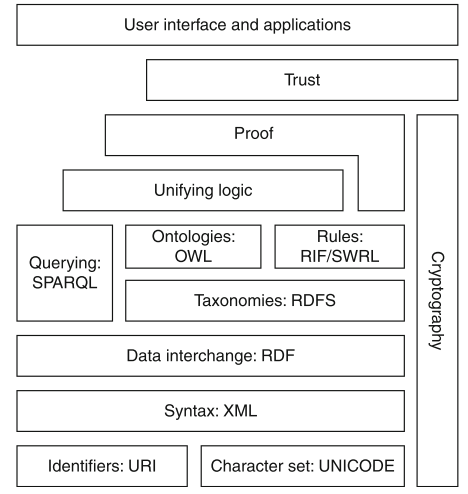}
	
	\caption{Layered Technology stack of the Semantic Web}
	\label{fig:layeres-technology-stack-of-semantic-web}
\end{figure}

Figure~\ref{fig:layeres-technology-stack-of-semantic-web} depicts the technology stacks of the Semantic Web which is a layered diagram containing all major technologies used in the Semantic Web today. The foundation of the technology stack is the identifiers denoted in URIs and characters represented in Unicode. Then XML is selected as the main syntax form and on which RDF layer is placed. The RDF is a simple yet powerful knowledge representation in the form of triples. A RDF triple is composed of a subject, predicate and an object which are represented in URI form. Current RDF schema allows objects to be denoted as literal values instead URIs, however, subjects must be represented in URIs pointing to entities being mentioned. Triples in Semantic Web represents the basic element of information which are collectively organized under ontologies specified in OWL. In addition, RDFS can also be used to denote a simple taxonomy, however, it is not powerful as OWL when specifying complex relationships. Semantic Web also provides a querying language, SPARQL which is similar to the relational database querying language SQL, however, it is more expressive and has a number of constructs specific to the Semantic Web infrastructure. Semantic Web also contains the logic specified in Description Logic and also has its roots in Graph theory. However more importantly, covering all of the layers in the technology stack, the security layer enables the privacy and protection for the Semantic Web data. 

The collection of the aforementioned technology stack helps the Semantic Web to be applied in number of data intensive application areas. Furthermore, it is also supported by number of active communities, conferences (e.g., ISWC, ESWC) and W3C (and its working groups) which set the standards for the Semantic Web.

\section{Oil and Gas Industry}
\label{sec:oil-and-gas-industry}

As introduced briefly at the start, Oil and Gas industry is one of the most data intensive industries in the world. This has influenced a number of Oil and Gas companies to manage the massive data load using the Semantic Web technology stack. 

The Semantic Web applications in the Oil and Gas industry can be described under four main categories. Table~\ref{tab:the-categories-sw-oil-gas} describes these criteria where first three were initially proposed by \citet{Chum2009} and we have added the new category,data security, which focuses on the ways of improving the privacy and security of the data when applying the Semantic Web infrastructure. 

\begin{table}
	\caption{The categories where Semantic Web applications considered in the oil and gas industry\label{tab:the-categories-sw-oil-gas}}
	\begin{tabu}{X[-1.3]X}
		\toprule
		Category & Description \\
		\midrule
		Data standardization & This factor focus on how current oil and gas data can be standardized through a Semantic Web inspired solution\\
		&\\

		Data integration & This covers how Semantic Web can combine data coming from different processes in oil and gas industry and providing a unified view to analysts.\\
		&\\

		Data sharing and enhancement & This covers how Semantic Web technologies help to share data among users, applications, and other processes. And it further discuss the chances of enhancing the oil and gas data.\\
		&\\

		Data security & This will discuss security issues which can arise in Semantic Web enabled oil and gas industry.\\
		\bottomrule
	\end{tabu}
\end{table}

Although applications related to all these areas are not overly prevalent in the industry today, there are several case studies that focus on different aspects of the four mentioned categories in Table~\ref{tab:the-categories-sw-oil-gas}. Specifically, Chevron has carried out number of projects to investigate how Semantic Web can be beneficial to their oil exploration and management workflow. 

\citet{Soma} present the semantic web solution for real-time reservoir management system developed at Chevron. The approach uses the semantic web to access data efficiently and to enhance a consistent view of the information. The core of the system is relying on a meta-catalog which is a knowledgebase powered by semantic web technologies. The system first extracts information from the repositories and enhances the data using OWL inference before being uploaded to the knowledgebase. Furthermore, a SPARQL based querying function is also provided, so that information can be accessed more efficiently compared to searching the semi-structured information.

Fluor Corporation\textquoteright s Accelerating Deployment of ISO 15926 (ADI) project \citep{OnnoPaap2006,Paap2008} describes how RDF/OWL and ISO 159263 Part 4 Reference Data Library can be utilized to enhance the information available in the whole life cycle of a plant. The ISO 159263 Part 4 contains definitions of plant objects and is considered as a main resource in oil and gas industry. However, as this information is text based, its usability in computerized systems is very low. Therefore, the ADI project transforms this information into RDF/OWL form so that computer systems can understand and further enhance the data. In addition, the OWL form of the definitions also helps corporate level data sharing as well as update the information on demand. 

Similar to ADI project, Norwegian Daily Production Report (DPR) project also uses the ISO 15926 to standardize the production data reporting. DPR also receives number of benefits by applying the ISO standard based ontology. Mainly, as the production data reports use the computer understandable form of classification (i.e., ontology), the reports can be queried easily and even the data can be compared automatically supporting automated decision analysis. 

It has also become an important goal to integrate ontologies with the increase in ontology schema to organize oil field information. The Integrated Information Platform (IIP) project focuses on this special objective where a platform is designed to integrate ontologies and generate new ontologies. The IIP project currently contains the information from Petrotechnical Open Software Corporation (POSC) which provides number of ontology classes based on ISO 15926. Furthermore it is also important to increase the awareness of such projects, so that the ontology integration can be carried out broadly resulting a conformance within the oil and gas industry for data definitions.

The case studies discussed in this section illustrated number of different ways that Semantic Web is employed in the Oil and Gas industry. However, still there are many opportunities to utilize the Semantic Web concepts in Oil and Gas to improve the workflow and the quality. For instance, Semantic Web can be combined with Internet of Things (IoT) devices and other hardware accessories that are used to generate information in structured from with underlying metadata. Such information generation can enhance the information quality as well as supports the underlying information infrastructure. Furthermore, the need of human friendly systems to access these information has also become a serious need. As Semantic Web data is represented in triple form unlike natural text presented in a document (as in traditional approach), the accessibility of the data for human users has decreased. Due to this, middle-tier frameworks to transform SPARQL, triples, and ontology specifications to natural language and vice versa has become an important need.

\section{Military Technology}

With the invention of novel military technologies that range from weaponized mobile robots to armed predator drones, the amount of information produced is massive and grows exponentially with the number of instances placed in the battle. More importantly, all the information produced by these automated military units are important in making decisions on the battle. Military technologists are now very keen on Semantic Web to manage the growing data load and support the decision analysis phase during the battle. 

\begin{figure}
	\includegraphics[width=0.7\linewidth]{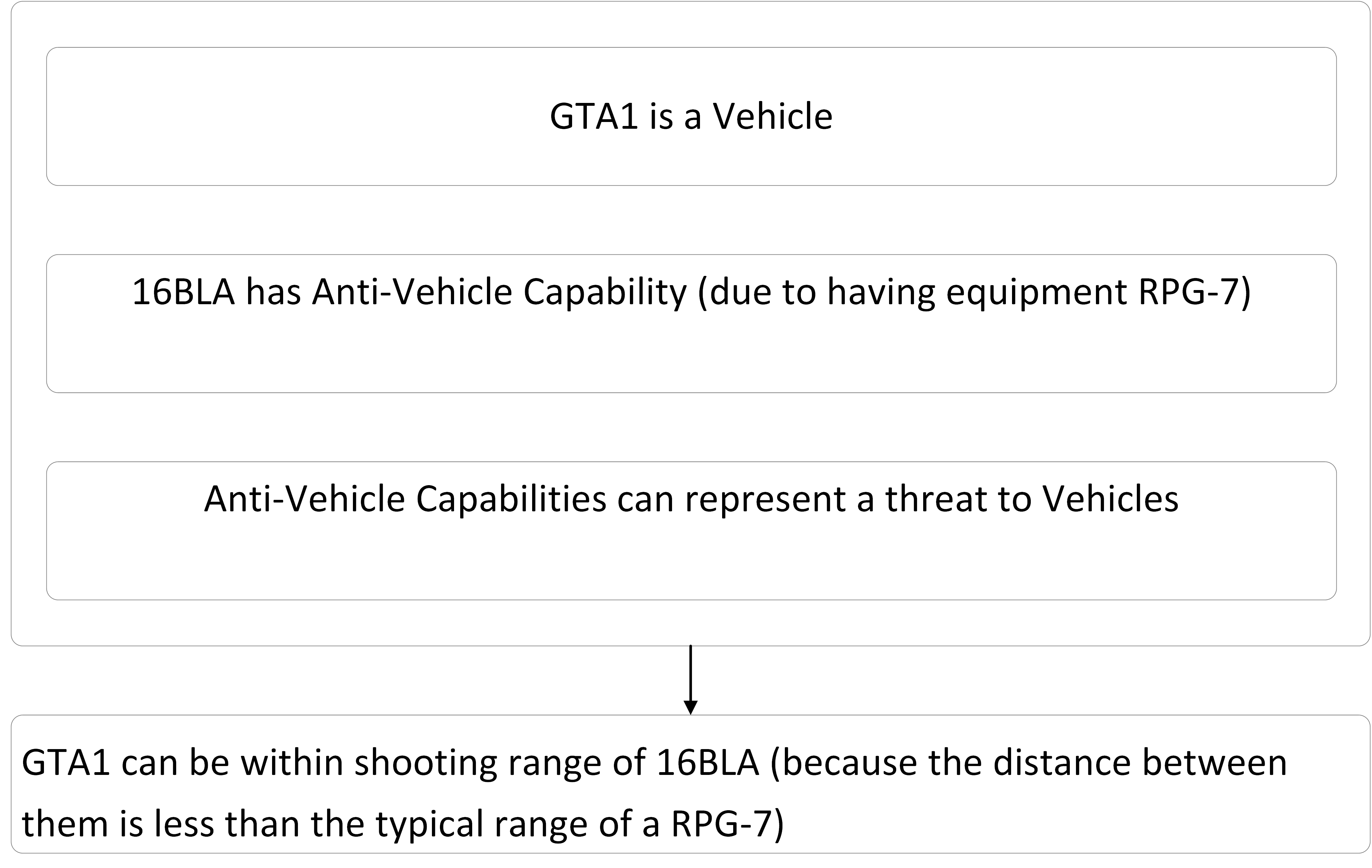}
	
	\caption{Military Information Integration Example\label{fig:military-information-integration}}

\end{figure}

\citet{Halvorsen2012} introduce a military information integration approach using Semantic Web technologies. Military systems that are placed in a battle or towards a common goal often need to share the information to carry out successful interconnected mission. It is also important that human agents or computerized agents involved in the battle get background information on the objects and environment of the mission is taken place. \citet{Halvorsen2012} utilize the RDF and the information presentation and serialization mechanism between agents and SPAQL as the communication protocol. The proposed communication protocol contains both the one-off SPARQL queries as well as streaming queries which offers agents to share information or get updates continuously on a certain task or change in the environment. In a case study presented by \citet{Halvorsen2012} shows how this Semantic Web based information integration system can be used for threat detection. The case study includes number of information sources including DBpedia which is an open domain Linked Data source. Using these information source, the system attempts to identify the threats nearby by reasoning over number of information units which are provided as the RDF triples and acquired through SPARQL queries. Figure~\ref{fig:military-information-integration} shows an example scenario where threat detection is taken place using the RDF triples and simple reasoning over the triples.

\begin{figure}
	\includegraphics[width=1\linewidth]{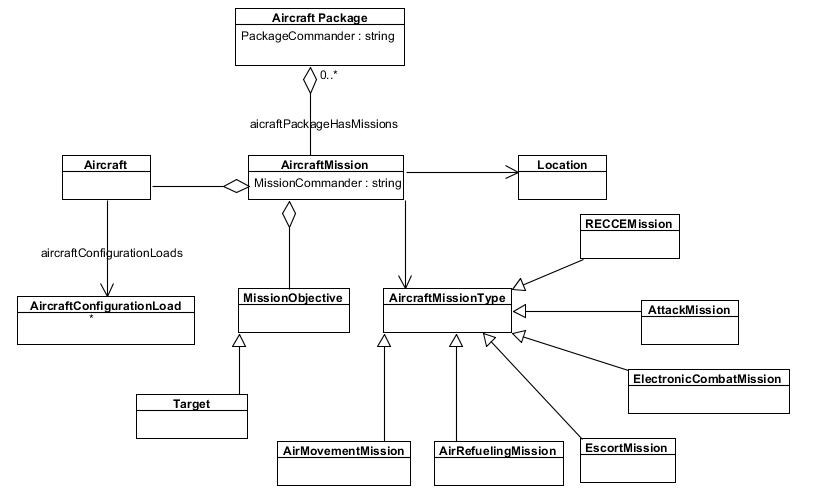}
	
	\caption{A portion of the aircraft mission ontology which is designed by \citet{frantz2005semantic} }
	\label{fig:aircraft-mission-ontology}
	
\end{figure}

The Military domain is also keen on applying the Semantic Web techniques to standardize the information that it generates. For instance, MilInfo \citep{Valente2005} is an ontology that contains the military information and is currently comprised of 188 concepts. The main benefit of developing MillInfo is that the information can be shared between the systems and information from different systems can be integrated to support the automated decision analysis. As described previously in the approach presented by \citet{Halvorsen2012}, the information required for a battle may reach the decision analysis system from different devices, soldiers and even from external systems. The availability of an ontology such as MillInfo makes the information integration process easier and more accurate. Similarly, numerous efforts have been taken to develop ontologies to conceptualize the battlefield information. Air Tasking Order (ATO) ontology \citep{frantz2005semantic} is used as the formal document to assign aircrafts to missions. Since ATO is a rich terminology, the corresponding ontology also contains number of classes which presents the formal definition of the knowledge. Figure~\ref{fig:aircraft-mission-ontology} illustrates the aircraft mission ontology which is a part of the ATO and it also utilizes another two ontologies developed as a part of ATO, aircraft and aircraft configuration load ontology. \citet{frantz2005semantic} also describe the process of building the ATO ontology which uses the OWL and has also carried out reasoning experiments which shows good potential in numerous military applications. 

Several other ontologies such as Tactics, Techniques, and Procedures ontology \citep{lacy2005experiences}; Air Mobility Command (AMC) ontology \citep{mulvehill2004act,mulvehill2005}; Computer Generated Forces (CGF) ontology collection \citep{lacy2005experiences} which is comprised of 4 different ontologies; and as well as Battle Management Language (BML) ontology \citep{turnitsa2006battle} also provide a formal conceptualization of the military information which can be used for machine-to-machine interaction for enhancing the mission effectiveness and enabling rapid situational awareness \citep{pulvermacher2005perspectives}.

The analysis shows that Semantic Web has number of applications in the military operations in order to make the safety as well as to achieve the goals in ease. However, there are also some opportunities to integrate Semantic Web to further improve the military operations as well as to come up with informed decisions. One of the areas that needs immediate attention is the transformation of information about soldiers to Semantic Web format (i.e., Linked Data). This has number of benefits as the data can be used to select soldiers who are ready to include in a particular battle as well as can also be used to keep track of the soldiers who are retired. The selection of soldiers for a particular battle can be extremely challenging when the mission is in a critical stage. Thus it may require selecting soldiers who have prior training on a similar terrain, fluency of a particular language, or even who have similar physical appearance so that they can reach to the enemy without being noticed. In such scenario it is important to record all the information of a particular soldier as well as an expressive querying mechanism to easily select limited number of candidates from a large group (e.g., selecting 5 most suitable soldiers from a 10000 group to execute a critical mission). Secondly, the retired soldiers (mostly high grade officers) have reasonable knowledge about the battle field as well as sensitive data, it is vital to keep track of these members in a constantly updated and linked information source which supports advanced querying mechanisms. The Semantic Web fulfills these gap with its graph storage and SPARQL querying which is expressive than traditional relational database querying such as SQL queries. 

AKTiveSA project \citep{Smart2007,Smart2007a} investigates the Semantic Web application in Military Operations Other Than War (MOOTW). MOOTW covers situations such as humanitarian relief, re-establishing normalcy in an area, noncombatant evacuation operation, and no-fly zone enforcement. Since a number of diverse agencies are connected in such scenarios, the MOOTW operational team has to deal with a large number of heterogeneous data sources. AKTiveSA project employs Semantic Web to improve situational awareness in such operations with a full stack of semantic technologies. In essence, the ontologies are used as the basis for knowledge conceptualization and to integrate information rapidly, while RDF store is used to store the information with a SPARQL endpoint which is provided to access information with high expressive power.

\section{e-Government}

The concept of government has evolved from the first simple form of government created by the Sumerians using clay tablets as a reporting tool to the 196 governments in the world today which operate based on complex and advanced technologies including Artificial Intelligence (AI). However, with the advancement of the government processes and amount of data produced in various departments, arises the need for scalable and enhanced data managing platforms.

Semantic Web addresses this growth of government data with the new concept of e-Government where process is largely based on IT related processes. \citet{Yang2006} present the semantic information portal and the semantic search algorithm for the e-government domain. The implementation of a semantic information portal has numerous benefits compared to the traditional information portal which is based on unstructured text. A semantic information portal can offer advanced search mechanisms as the information is organized on a domain ontology which defines the conceptual representation of the data. The underlying search mechanism can utilize the ontology to come up with search results which are more specific to the query instead of a text search as carried out in traditional information portals. Furthermore, a semantic information portal supports the decentralization of information by defaults unlike a centralized traditional information portal. This decentralization is a major advantage when scaling the information portal to keep massive amount of information and serving exponentially growing user base. In addition, semantic information portal also supports the automation of the services as the information can be understood by the computer program as the information is unambigiously associated with the underlying metadata. Therefore, applications such as decision analysis systems can be developed using the information recorded in the semantic information portal which enables the e-Government services to operate increased efficiency. \citet{Yang2006} also implement the semantic search algorithm to retrieve results for a given query. Instead of a traditional search approach which with text search, semantic search algorithm can also associate the metadata associated with information and can hence can achieve a deeper search on the units of information. 

Building ontologies for the e-Government is also at the center of the semantic web based e-Government applications. \citet{Haav2011} describes the process of developing domain ontologies targeting semantic annotation of e-Government data services. \citet{Haav2011} introduces a process comprised of seven steps to build domain ontologies for e-Government which starts with the collection of concepts and then follows an iterative development. The advantage of such specified ontology design is that it supports the changes with minimum effort and the process can be reused in a number of scenarios with minimum modifications. Several other researchers \citep{4221189,Hinkelmann2010,Gomez-Perez2006,Han2010,Salhofer2009,Sanati2012} also discuss how ontologies can be successfully utilized for e-Government applications. The core advantage as pointed out by these researchers is that an ontology offers the conceptualization of information which was earlier recorded as unstructured text with no associated meaning. 

\begin{figure}
	\includegraphics[width=1\linewidth]{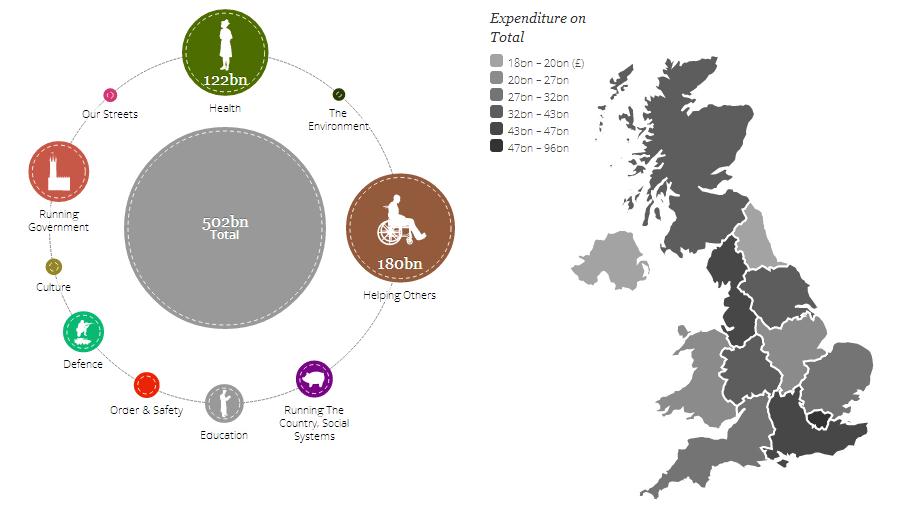}
	
	\caption{A portion of the UK government open data visualization to analyze the money distribution among different departments}
	\label{fig:uk-open-data-where-money-goes}
\end{figure}

In addition to the ontologies and the applications, governments are now voluntarily engaged in the open data initiatives where government data is published to the public to be accessed freely.. Majority of this data is statistical data which describes the performance factors as well as some figures related to the government operations. Therefore, RDF Data Cube vocabulary which is introduced to model statistical data has become the defacto data model for open data published by governments. For instance, United Kingdom has made major effort towards releasing its data as open linked data utilizing the RDF Data Cube form accessible via 19 publicly accessible APIs \citep{Tom2009}. This approach increases the transparency of the government as well as open new research opportunities to increase the quality of the government based on the historical data. One of the significant example is the analysis and visualization carried out on how UK government spending money. Figure~\ref{fig:uk-open-data-where-money-goes} shows a portion of this visualization where distribution of money among different governments is presented in a graph view.

This section discussed a number of different ways in which the Semantic Web is applied in e-Governence in order to increase the accessibility of the data as well as to open new opportunities for research and as a result the quality of the governance is improved. The approaches mentioned here mainly focused on ontologies, semantic services, data representation, and visualization. However, the structured form of the government data also generates a number of other benefits towards intelligent use of this data for predictive analysis and future resource planning. Although such intelligent analysis can be carried out at the government level, it will become even more important to look at this from the global perspective which enables us to get ready for any upcoming global challenge. In addition, the core factor behind such global analysis is to improve the linkage of government datasets. In essence, we also need to focus on automated models to align and link government data thus generating a massively interlinked government Linked Data cloud which provides a view on how different governments perform in the globalized world.

\section{e-Business and e-Commerce}

E-Business and e-Commerce are also key areas that handle massive amounts of data and utilize this data to make important decisions which have substantial impact on the future of the organization. Although data is central these applications, a large proportion is still kept as unstructured data. However in recent times, there has also a steady trend towards the use of Semantic Web in business and commerce areas.. In this section we discuss how e-Business and e-Commerce employ Semantic Web to to implement smart solutions that reduce the cost and increase efficiently leading to higher overrall benefits.

One of the key requirements to enhance the business process with Semantic Web is to come up with ontologies that can conceptualize the business workflow. \citet{hepp2008goodrelations} introduces the GoodRelations ontology which can act as a core requirement for a successful Semantic Web based e-commerce platform. GoodRelations ontology can be used to model e-commerce information about products, prices, offers, and many other concepts that are frequently used in the commercial environment. This extensive representation of e-commerce entities makes the GoodRelations ontology one of the fundamental tool in applying Semantic Web in e-commerce domain \citep{ashraf2011open,hepp2012web,niknam2013semantic}. 

\citet{akanbi2014lb2co} presents the LB2CO, an integrated e-Commerce ontology framework. The LB2CO introduces two types of agents, namely, search agent and ontology agent to increase the usability of the e-Commerce applications. The search agent is the main communication point to the consumer and it is capable of analyzing metadata associated with the products. The ontology agents manages the information where products are described ontologically, so that search agent can utilize the information and answer queries efficiently. \citeauthor{akanbi2014lb2co} also evaluates the system with a case study where a prototype is built based on the ontological framework. The approach presents promising results where the consumers queries can be appropriately processed using the semantics embedded in the query and also there is a wide range of future enhancements for such ontological framework based e-Commerce platform. 

Matchmaking in e-Commerce is yet another area which has widely benefited from Semantic Web. Matchmaking in the e-Commerce context corresponds to narrowing down the search space of product or service offers in order to select the best compatible one available. Hence, matchmaking systems need to deal with a number of different factors which can be numerical scales as well as categorical variables which define requirements which ultimately need to be aligned with the offer. This offers a good opportunity to employ Semantic Web as it can work on the structured level of data and have the flexibility of defining the data with semantics which is very important in the matchmaking process. \citet{paolucci2003toward} describe the implementation of DAML-S Matchmaker which is comprised of ontologies, DAML-OIL reasoner, and a matching engine. DAML-S Matchmaker is capable of automating business processes by interacting with the web services by utilizing the DAML annotated data. For instance, a user can automate the entire process of planning a trip by initiating the process with a date. DAML-S Matchmaker is then capable enough to interact with e-commerce applications including airline services, car rentals, and even accommodation services and complete the rest of the process for the specified date. The ontology reasoning and other semantic technologies are cornerstones of this process as they helps the matchmaker to come up with possible solutions for the user needs. \citet{trastour2002semantic} focus on utilizing DAML+OIL to develop an expressive and flexible web service description language to support business-to-business e-commerce domain. \citeauthor{trastour2002semantic}'s proposed system contains the entire lifecycle of e-commerce which includes matchmaking, negotiation, contract formation, and contract fulfillment. The systems is demonstrated using a prototype matchmaking system that uses FaCT++ (OWL DL Reasoner) and operates on service descriptions in DAML+OIL.

\begin{figure}
	\includegraphics[width=1\linewidth]{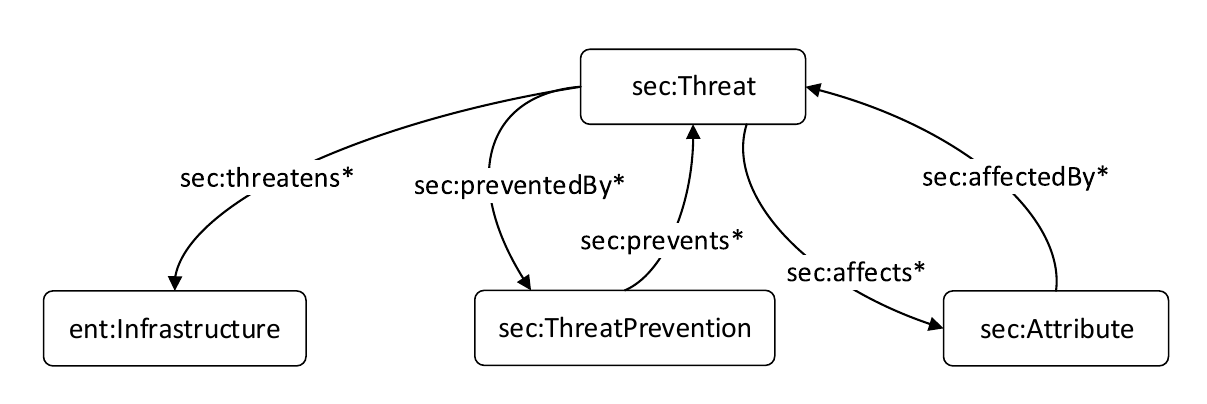}
	
	\caption{A portion of the security ontology introduced by \citet{Ekelhart2007}}
	\label{fig:portion-of-security-ontology}
\end{figure}

The security of e-Commerce applications based on Semantic Web has also become a key area to focus as the application of Semantic Web can significantly change the information management and sharing strategy. . A detailed discussion on security issues that can be present in the Semantic Web based e-Commerce can be found in analysis carried out by \citet{Ekelhart2007}. According to \citet{Ekelhart2007}, one of the extremely necessary need in Semantic Web based e-commerce is to protect the integrity and availability of the ontologies. \citet{Ekelhart2007} propose XML access control as well as introduce the security ontology that helps to model security issues in the e-commerce domain. Fig.~\ref{fig:portion-of-security-ontology} illustrates a portion of the security ontology introduced by \citeauthor{Ekelhart2007}. The figure indicates only the threat section of the security ontology which is composed of three main sub-ontologies, namely, \textit{Threat}, \textit{Threat Prevention}, and \textit{Infrastructure}.

This section illustrated how Semantic Web can be utilized in e-Business and e-Commerce domains to conceptualize data, build applications, as well as to improve security. However, there is an obvious need for movement towards Semantic Web based data representation within business organizations. One of the major factors that influence this movement is the difficulty in employing unstructured text in the decision making process. In essence, unstructured text is ambiguous and finding the correct information using token matching takes time and manual effort in a large text corpus related to the business processes. On the other hand, Semantic Web triples encoded with semantics and schema specifications are much more precise and in addition the triples can be queried directly hence eliminating the need for errorenous for token-matching based search.

\section{e-Health and Life Sciences}

The Entire health and life science industry is now in the transition to the e-Health movement where patient data will be kept and analyzed electronically (possible using AI in the future) rather than paper based documents. However, to develop solutions that work efficiently with electronic patient data that can cater for millions of users, it needs to have scalable and queryable forms of data representation. In addition, to support integration of AI in the future for the automation of the decision making process, this data representation should also support holding metadata which describes the semantics of the data supporting automated systems to understand the data. 

\citet{Krummenacher2007} describes the process of building the European Patient Summary (EPS) infrastructure based on Semantic Web technologies. The EPS project focuses on building a European eHealth infrastructure that is capable of providing a data accessing and processing platform for primary clinical data stored in eHealth applications across the European healthcare delivery network. This aim require EPS to develop a scalable infrastructure that can integrate information in an efficient manner. The EPS project, therefore has selected Semantic Web that has already shown the capability of information integration in a massive scale.

Semantic Web has also been employed by \citet{Podgorelec2007} to integrate and store diagnostic data of Mitral Valve Prolapse (MVP) syndrome. The architecture of the system is based on inter-connected modules where RDF provider plays the main role of data acquisition from different sources and integrate them to increase the accessibility of the data. The data integration framework can also interface with a web portal where the data can be easily accessed and searched. Specifically, searching of the data which is associated with metadata can significantly decrease the ambiguity level of the search results.. 

In addition to the RDF data storage, the ontologies also provide a number of benefits to the eHealth application by presenting a conceptualization of the domain knowledge. For instance, \citet{Kim2007} present the design of the knowledge base for heart disease detection. The knowledge base encodes knowledge acquired from ECG and other information sources using the semantic web as the underlying technology and more specifically focusing on ontologies to conceptually organize the knowledge. The ECG ontology which is introduced in the research provides the conceptual overview of the wave properties of ECG such as polarity, lasting period, amplitude, and shape of wave. This type of conceptualization can ultimately support the automated decision analysis based on the knowledge base and can vastly reduce the workload of the medical practitioners in the future. Although, ontologies can contribute to eHealth in a positively beneficial way compared to a simple taxonomy, mismatches can also occur which could lead underlying knowledge infrastructure into confusions. The research proposed by \citet{Ganguly2008} focusrd on addressing the interoperability issues in ontology based eHealth applications by resolving mismatches between the conceptual hierarchies. 

The case studies discussed above shows how Semantic Web has contributed to eHealth by leveraging the information accessibility and integration. Some workshops such as Semantic Web for Life Science (SW4LS) are specifically devoted to discuss the impact of the Semantic Web in the eHealth domain. However, still number of research challenges and opportunities exist in integration of Semantic Web with eHealth domain. Firstly, a movement towards rapid conceptualization of medical and health terminologies is essential. Although, Human Phenotype Ontology (HPO) \citep{kohler2013human}, Systemized Nomenclature of Medicine Clinical Terms (SNOMED- CT) \citep{de2011systematized} and several other ontologies have contributed to come up with a detailed specification of conceptualization, there still remains large amount of medical and clinical knowledge that is not included in the ontologies. In addition, certain standards and protocols need to be devised before introducing Semantic Web as a consensus data storage (i.e. in triple stores) medium for eHealth applications as data usage in eHealth domain has number of ethical and security constraints.

\section{Multimedia and e-Culture}

Netflix, Spotify, Vevo and other related entertainment services have transformed the entire way we consume multimedia content, boosting the industry with advanced technologies including AI based recommendation algorithms. However, with approximately 2500 movies and 1 million songs released to the public annually, the metadata which carries important information about these releases such as producers, actors, musicians, and even recording companies are also growing at lightning speed. The multimedia services need this metadata to be able to offer quality services to their users will also need to be able to handle the growing data load. 

\citet{4061414} propose an image retrieval approach utilizing ontology-guided reasoning. The proposed approach uses the ontology to identify the semantics associated with user queries and Bayesian Network (BN) to calculate the relevance of the images to the initial query. The ontology in this approach plays a significant role on conceptualizing the user query by enhancing it with subclasses, superclass, and the equivalence classes. This semantics-enabled approach significantly deviates from the traditional approach which focuses on mapping the user query keywords to the images with no underlying semantics or conceptualization of the query terms. Ontology utilization in image retrieval is investigated further by \citet{Wang2007}, introducing the multimodality ontology to annotate and retrieve animal images. The multimodality ontology is comprised of an animal domain ontology, textual description ontology, and a visual description ontology. All these three ontologies provide conceptualization of the animal information and are utilized to annotate and retrieve animal images. Such an approach can further fine tune and enhance the image retrieval systems, however, it is also important to focus on how these ontologies can align with each other resolving any mismatches. 

The aforementioned two approaches focused on Semantic Web ontologies to retrieve image ontologies. However, \citet{Liu2010} focus on reverse engineering the process by generating the image ontology from the images. The core of the research resides on building the ontology using the SIFT feature vector which is used as properties. In addition. Color, texture, and shape are also used to build the ontology. The importance of building the image ontology in this way is that retrieval of images can be fine-tuned using the feature vector which is more specific in image annotation. 

The Semantic Web concepts are also applied in the video annotation and retrieval to increase the accuracy by associating the underlying semantics. For instance, \citet{Ballan2010} present a framework to annotate and retrieve videos using ontologies and rules expressed in Semantic Web Rule Language (SWRL). The annotation process proposed in this research automatically annotates events in a video with semantic annotations using a rule based approach. The system learns these rules using rules specified in SWRL which can also be used to reasoning which makes the \citeauthor{Ballan2010}'s approach scalable in the video annotation domain. 

This section introduced existing methods that Semantic Web is employed in multimedia and eCulture domain. However, there is also an opportunity to use Semantic Web to improve the user interaction by improved suggestion algorithms that work based on Linked Data triples. For example, a user may be watching a particular set of movies or TV series due to a reason such as a common actor/actress, genre, or even the filming location. One of the future tasks that Semantic Web should focus on is that associating this type of information with multimedia content as Linked Data, and build a user model which can be represented as a Linked Data graph. Once such model is available, machine learning algorithms can consume this information as features and can come up with predictive data analytic models to provide suggestions for a user. In addition, the production companies can also utilize Semantic Web enabled multimedia data platforms and machine learning models to decide on what themes they should focus in the future and which actors/actress they should use for the upcoming movies.

\section{Geo Information systems}

Geo Information Systems (GISs) rarely act individually \textemdash{} in general they are interconnected, and data from different sources have to interact in order to achieve an objective. This data interaction becomes extremely challenging with the massively increasing amount of data coming from multiple geo information systems. In some cases when adapting an already available geo information systems for a different domain, terminology matching is needed which ultimately requires data re-purposing. Another key feature required from existing geo information systems is querying of the integrated interconnected data. Semantic Web as a single technology stack offers unique solution for the aforementioned key challenges in Geo information systems. 

Semantic Web enabled ontologies have become a key component in geospatial data conceptualization. GeoNames ontology derives geospatial information from the GEOnet Names Server (GNS) and the Geographic Names Information System (GNIS) and presents the data in an ontology that can be processed by machine. One of the key advantages in using ontologies in GIS is that they provide an opportunity to carry out required reasoning which enables GIS to derive new information as well as better analyze the existing information. In addition, the ontologies also support the integration of information from different GISs to build a scalable geographical knowledge infrastructure. To support this information integration using ontologies, a number of researches are now investigating the geospatial ontology matching as a key task in GIS development \citep{Delgado2013,Du2013,Varanka2016}.

In addition to the data conceptualization, there is a clear surge in Semantic Web vocabularies for Geospatial data. For instance, GeoSPARQL \citep{Battle2012} focuses on designing a SPARQL standard which is more appropriate for geospatial Semantic Web data. GeoSPARQL extends the current SPARQL formalism by providing topological SPARQL extension functions and Rule Interchange Format (RIF) Core inference rules. These additions become essential when quantitative reasoning and query interpretation using Geospatial data. 

In addition to GeoSPARQL, number of vocabularies such as stRDF, Geom are also implemented focusing on Geospatial Semantic Web. Interestingly, these new vocabularies are also integrated into a number of commercial and open source Semantic Web frameworks (e.g., Apache Marmotta \citep{Apache2017}, Parliament \citep{kolas2009efficient}) to achieve a scalable and intelligent Geospatial Semantic Web. 

\citet{ODea2005} describe the Semantic Information Demonstration Environment (SIDE) project which is designed to explore the possibilities of using Semantic Web in the Geospatial analysis services. The SIDE project allows users to query heterogeneous data sources with the single query client which is provided with a query interface which can be used to formulate the query. The query is then executed on data sources including Relational Database Management Systems (RDBMS), web data sources, and RDF instances which are stored locally. In addition, the SIDE employs acore ontology, Description Logic (DL) reasoner, and a rule based reasoner. Although SIDE is not available to access, the proposed system offers promising insight into the usage of semantic web based technologies to make the geospatial information widely accessible.

This section showed the importance and usage of Semantic Web in geo information systems. The future work in Semantic Web frameworks for geo information systems should focus more on enriching raw data with additional semantics. In addition, there is a definite need for a systematic method of designing ontologies which can conceptualize the the rich knowledge essential for Semantic Web frameworks.

\section{Investigative and Digital Journalism}
\label{sec:investigative-journalism}

Investigative Journalism is the new face of the journalism that uses deeper level research on a specific case and compiles a report in with information that would otherwise be either unattainable or difficult to attain manually by a journalist. The wide ranging availability of data and a smart and focused approach makes the derived information much more richer and insightful than the information that can be compiled by a journalist usually working under extreme time constraint. 

Although the information is important and the approach is novel from the journalism perspective, the presentation mode still contains inherent accessibility issues. For instance, an investigative journalist can reveal a serious crime and compile a report mentioning a number of previously hidden reports/communication while making them public. However, the full benefit of these important documents is usually not reaped as these documents are based on unstructured text which need significant amount of effort to read, understand, and link entities. Semantic Web comes into play a significant role when investigative journalism faces such hurdles of presenting the information in an accessible manner. 

\begin{figure}
	\subfloat[Links between different entities in the Linked Leaks dataset as reported by \citet{Kiryakov2016} ]{\includegraphics[width=0.45\linewidth]{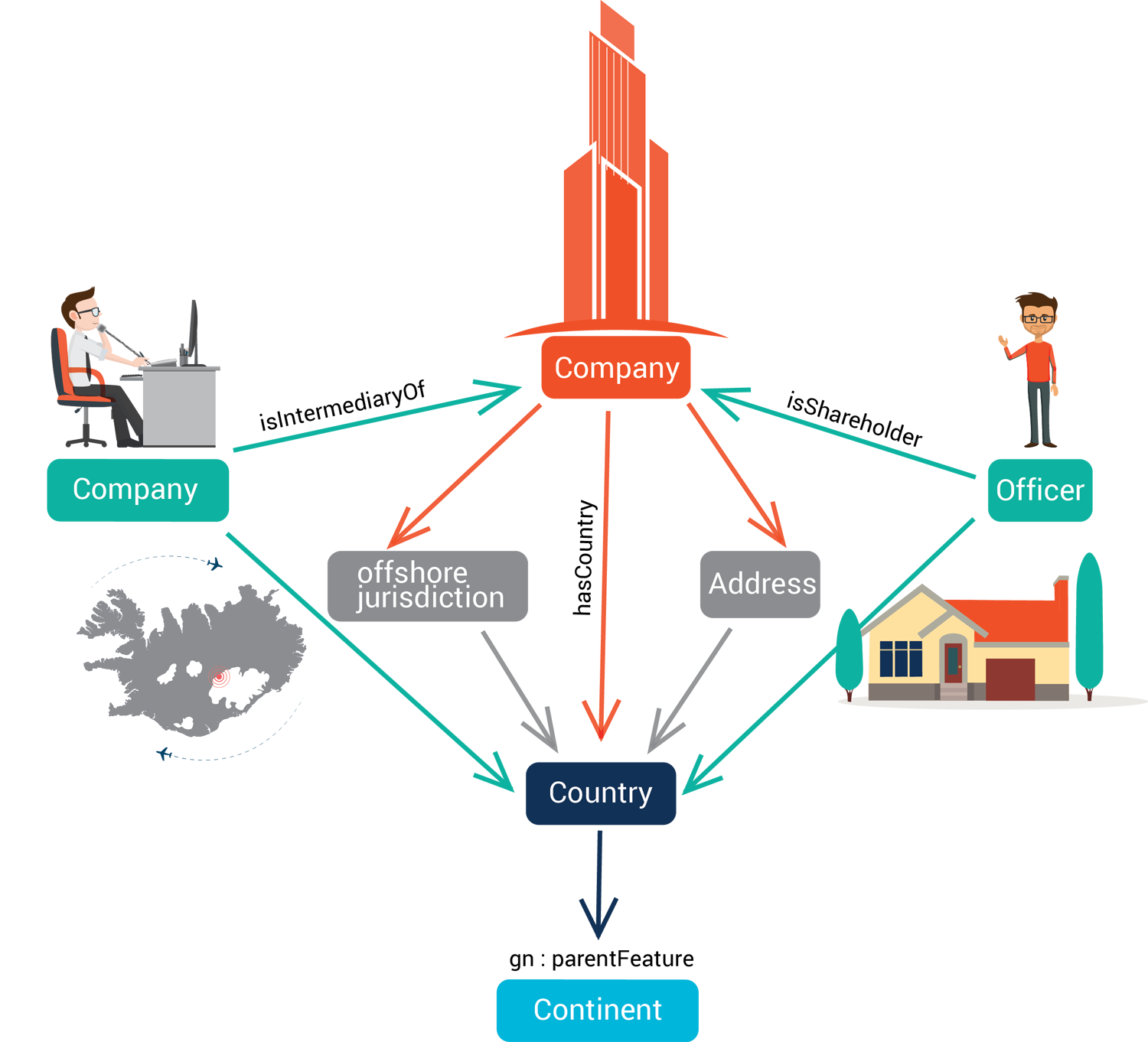}
		
	}
	
	\subfloat[Class relationships in Linked Leaks dataset]{\includegraphics[width=0.45\linewidth]{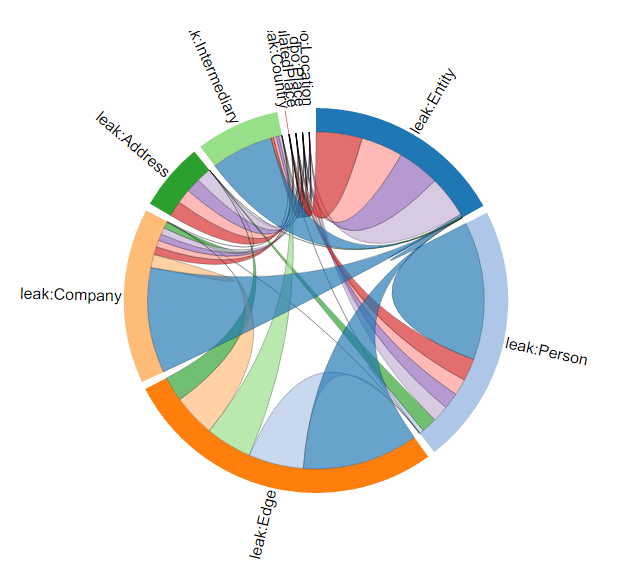}

	}
	
	\caption{Links between different entities in the Linked Leaks dataset}
	\label{fig:linked-leaks-example}
	
\end{figure}

Panama Papers is the largest known document list that investigative journalists ever worked on. The document list \textemdash{} which belongs to the the Panamanian law firm and corporate service provider Mossack Fonseca \textemdash{} was leaked by an anonymous group. The information encoded in this document list contained how individuals and organizations evade taxes and other sanctions. Although the document list contained very important information that can generate number of interesting articles, it was a great challenge to analyze and link the natural language information. For example, a mention of a person appears in number of different locations in the document list and sometimes referred as his/her associated organization (e.g., the company he/she owns). Meanwhile, Ontotext \textemdash{} a company that focuses on Linked Data tools \textemdash{} transformed the document list to Linked Data form and released the new dataset which was introduced as ``Linked Leaks dataset'' \citep{Kiryakov2016}. The Linked Leaks dataset contains the data in the structured form and can be queried using SPARQL. These two features make the Panama Papers information easily accessible and also automatically include all the links between different entities. Figure~\ref{fig:linked-leaks-example} shows an example of the linked entities which is adapted from the technical report published by \citet{Kiryakov2016} and class relationships in Linked Leaks dataset. Listing~\ref{lst:linked-leaks-sparql} shows a SPARQL query that can be executed on the Linked Leaks dataset to find the politicians mentioned in the dataset.

\begin{lstlisting}[caption={SPARQL query that can be executed to find the politicians mentioned in the Linked Leaks dataset},captionpos=b,label={lst:linked-leaks-sparql}]
PREFIX onto: <http://www.ontotext.com/> 
PREFIX leaks: <http://data.ontotext.com/resource/leaks/> 
PREFIX dbr: <http://dbpedia.org/resource/> 
PREFIX dbo: <http://dbpedia.org/ontology/> 
PREFIX rdfs: <http://www.w3.org/2000/01/rdf-schema#> 
PREFIX rdf: <http://www.w3.org/1999/02/22-rdf-syntax-ns#>
SELECT DISTINCT ?leak_entity ?dbp_entity ?country_name 
FROM onto:disable-sameAs 
{ 
	?leak_entity leaks:mapped-entity ?dbp_entity.
	?dbp_entity ?p dbr:Politician.
	?dbp_entity ?p2 ?country .     
	?country a dbo:Country .     
	?country rdfs:label ?country_name .     
	FILTER(lang(?country_name) = "en") 
} ORDER BY ?country_name 
\end{lstlisting}

\citet{Fern2006} describe the NEWS (News Engine Web Services) project which introduces the Semantic Web to the news agencies. The NEWS project introduces a semantic annotation component to associate news items with metadata and an intelligence information retrieval component. The NEWS project also defines an ontology which is expressed as a lightweight RDFS ontology which is comprised of three modules, namely, a categorization module, a management metadata module, and a content annotation module. The categorization module supports the automatic news categorization by providing a vocabulary which contains 12 different top level subjects and contains 1300 categories. The management metadata module annotates the news items with information such as authorship, news item priority, and media type. The content annotation module is also focused on annotating news items with basic information.

Trump World Data \citep{Atanas2017} is another Linked Data resource that is currently subjected to very controversial discussions. The dataset contains the public declarations and investigations (carried out by journalists) of Donald J. Trump. Since Trump has significant amount of corporate connections which include a hotel in Azerbaijan and a poker company in Las Vegas. The dataset was initially released as tables and further analysis which can reveal deep relations was resource expensive on such dataset. However, Ontotext. a company which focuses on Linked Data, converted this dataset into RDF form and with a detailed ontology. Furthermore, the dataset was linked to open domain Linked Data resources such as DBpedia which increased both the value and validity of the data.

This section explained a number of investigative and digital journalism approaches which use Semantic Web concepts to increase the usability of the data. However, there is also a large open space to try different Semantic Web approaches in this domain. Specifically, there is a need for a systematic ontology design method for digital and investigative journalism. This will provide a unique and accessible modeling experience for the journalism, so that the data can be organized more clearly and consistently across different news agencies.

\section{Semantic Web Tomorrow}
\label{sec:semantic-web-tomorrow}

The previous sections discussed eight different application domains where Semantic Web is already applied with numerous benefits. This section focuses on presenting our view on future application areas of Semantic Web and some initial work carried out in these areas.

\subsection{Linking the Unlinked Crimes}

The famous quote, ``There is nothing new under the sun. It has all been done before'', is credited to the fictional private detective created by British author Sir Arthur Conan Doyle \citep{doyle2013sherlock}. The core idea of this quote is that every crime that takes place in present or in future has some link through similar characteristics to a previous one that has already taken place. Hence identifying these characteristics and building a linked cloud of crimes helps help us identify the nexus between cases and which would enhance and increase the effciency of crime resolution.. Semantic Web data infrastructure (i.e., Linked Data) supports this kind of data linking capability which can be directly used in criminal investigations. Although some research is already being carried out in similar areas like human trafficking \citep{Szekely2015} utilizing Semantic Web technologies, criminology remains largely untouched area for Semantic Web researchers. In particular, there is a clear shortage of ontologies developed (the only existing crime ontology we found is Italian crime ontology \citep{asaro2003domain}) to conceptualize the criminology domain and no effort has been taken to map unstructured text based detailed crime reports to structured form (i.e., triples) using Natural Language Processing (NLP) technologies including scalable relation extraction approaches such as OpenIE \citep{Etzioni2008,Mausam2012}, entity disambiguation, feature based machine learning and deep learning approaches.

\subsection{Connecting the Dots in Paintings}

One of the predominant skills of humans is being able to express a real or conceptual world artistically by way of paintings. Thus it has become one of the major areas which follows a number of styles (e.g., Western, far Eastern, African, Asian) using diverse set of media (e.g., oil, pastel, ink, acrylic), and influenced by number of artists (e.g., Pablo Picasso, Leonardo da Vinci, Rembrandt van Rijn). However, the conceptualization of this information (i.e., in the form of an extended ontology) as well as organization of this information (i.e., in the form of Linked Data) has still not been taken into consideration. For example, such conceptualization can pave a way to advanced as well as more accurate search function for paintings. For example, if a user wants to search portraits from Pablo Picasso during 1903 to 1908 period, then the only available method currently exist is a text based search using the information as keywords (e.g, Pablo Picasso painting 1903 to 1908). The accuracy of this search depends on the way the search engine understands the query as well as the available information. However, by forming a SPARQL query on a triple store which contains the information about paintings, a user can retrieve accurate information in structured form and reduces the burden of analyzing the web documents as in traditional search engine.

\subsection{Planning before reaping the harvest}

The modern agriculture is a science and unlike the traditional approaches, it currently employs number of complex scientific processes in order to reach the optimal harvest. However, one of the significant gaps that exists in agriculture today is the utilization of information in the automated decision analysis process. Since automated decision analysis is a computational process which needs information in a way that computers can understand, there is a clear need to transform unstructured agricultural information into a structured form before implementing automated decision analysis systems. This is a good opportunity to employ Semantic Web as it is designed specifically to prepare the information in order to make them computer understandable.

\subsection{Building the Tower of Babel again}

Some work has been already done in the area of linguistics and language processing which uses the Semantic Web. However, given that linguistics is a very advanced area, the contributions are not significant to state that linguistics has reasonably used the Semantic Web. Hence, we have included the linguistics and language processing as an area that Semantic Web should focus more in the future. The work carried out in the area include lexicalizing triples \citep{perera2016realtext}, generating triples using unstructured text \citep{perera2016kiwilod}, SPARQL query verbalization \citep{ell2012spartiqulation}, and transforming natural language questions to SPARQL queries \citep{kaufmann2006querix}. However, there is no significant effort being made on building relationship between multiple languages using the interlinked Semantic Web triples (i.e., as a Linked Data cloud). For instance, languages that belong to one family shares a number of similarities in both lexical and semantic aspects. Identifying and connecting these similarities using structured form can form a good foundation for language translations tasks as well as some other deep language processing tasks. In addition, we also need a detailed conceptualization of language families in the form of an extensive ontology which will also support the aforementioned task as well as to understand the science behind the languages more clearly.

\subsection{Connected Fashions}

The fashion and textile industry is embracing the digital world in a very rapid pace. The information which is related to the industry has become a cornerstone in this new transformation. For instance, the new trends in the fashion industry originated from different corners in the world, by different designers, promoted by different groups, and target different social groups. The information units which is related to a particular fashion is extremely important for business entities to make decisions, identify regions for promoting them, and even to design advertisements. Semantic Web offers the exact solution that needs to address this gap using interlinked information infrastructure and conceptualizations of knowledge in the form of ontologies.

\subsection{Maritime Web}

Maritime information systems manage vast amount of information ranging from geo spatial information to vessel tracking information. Every bit of information is important as the future planning significantly depend on the provided information. However, the current maritime information management systems work in isolation and and use significant amount of human hours during the decision making process. This is mainly due to the fact that the information is not stored with associated semantics, so that computers cannot understand the information without human intervention. This opens a clear opportunity for Semantic Web technologies which are designed to make the information understandable for the computers. In particular, the future maritime information management should focus on four different aspects: conceptualization of the information, extracting structured information from the unstructured information sources, information integration within different maritime service providers by linking the structured information, and integrating Semantic Web enabled geo spatial information. These four aspects lead the maritime information systems towards a Semantic Web powered maritime web.

\subsection{From Connected Drive to Linked drive}

Intelligent driver assistance systems such as ConnectedDrive, UConnect, and Entune, transform the automotive industry to the next dimension by offering a range of services which process vast amount of information. This information is structured as they are produced mostly by hardware systems or pulled from external web services. This makes these systems a step closer to integration of Semantic Web, however, this information is not represented in an interlinked form. The interlinking of information can consider ways of linking of the information produced by automobile to other external services as well as interlinking the information within different driver assistance systems. The latter approach is a timely need as such information integration between the systems is a way forward to collective intelligence in the automotive industry to make the self-driving cars with a high precision. Furthermore, the integration of Semantic Web in automotive industry will also support a range of benefits which will inherit from technologies that power up the Semantic Web. For example, the voice enabled driver assistance systems will be able to use the SPARQL as a information querying mechanism which will combine the expressive power of SPARQL and communication power of natural language.

\subsection{Interpreting Annual Reports}

An annual report presents a summary of information in natural language. The summary of the information illustrates the current status of an organization and its future plans, hence such reports contain critical information in a condensed form. However, the information is presented in natural language which required more processing time than structured information. This is because a reader has to read the whole report to get a clear overview of the information and how the information links to previous reports or to another information source. On the other hand, interpreting the information encoded in natural language is difficult and onerous during the decision analysis process based on annual reports which is required by financial advisers. These drawbacks can be successfully addressed by Semantic Web which focuses on linking information and associating data with semantics. Some initial work on transforming annual reports into Semantic Web format can be found in the recent work carried out by \citet{perera2016kiwilod}. However, further research on extracting triples from reports, linking them, and organizing the triples under a rich ontology schema is essential in order to apply Semantic Web to interpret reports.

\section{Conclusion}
\label{sec:conclusio}

This paper discussed the applications of Semantic Web in eight different genre of industries, including data intensive industries such as oil and gas exploration to trending industries such as investigative journalism. From the discussions, we illustrated that Semantic Web has contributed significantly to improve the state-of-the-art in these industries through the semantic web technology stack. However, we also noticed some of the gaps that need to be addressed in these industries, and they are also reported at the end of each section.

In addition to the current applications of Semantic Web, in Section~\ref{sec:semantic-web-tomorrow} we introduced another eight industries that we think will reap the benefits of the Semantic Web in the future. Although the list is not complete as Semantic Web will play a role in almost every industry on the earth in the future, however, we only focused on industries where direct and some fast integration of Semantic Web is feasible. 


\end{document}